# Cross-Machine Anomaly Detection Leveraging Pre-trained Time-series Model

Yangmeng Li[a]*, Kei Sano[c], Toshihiro Kitao[c], Ryoji Anzaki[c], Yukiya Saitoh[c], Hironori Moki[c], Dragan Djurdjanovic[ab]

[a] Program of Operations Research and Industrial Engineering, The University of Texas at Austin, 204 E. Dean Keeton Street, Austin, TX 78712-1139, USA.

[b] Walker Department of Mechanical Engineering, The University of Texas at Austin, 204 E. Dean Keeton Street, Austin, TX 78712-1139, USA.

[c] Equipment Intelligence & App R&D Department, Tokyo Electron Ltd., Daido Seimei Sapporo Building, 1-3 Kita 3-jo Nishi, Chuo-ku, Sapporo, Hokkaido 060-0003, Japan.

## Abstract

Achieving resilient and high-quality manufacturing requires reliable data-driven anomaly detection methods that are capable of addressing differences in behaviors among different individual machines which are nominally the same and are executing the same processes. To address the problem of detecting anomalies in a machine using sensory data gathered from different individual machines executing the same procedure, this paper proposes a cross-machine time-series anomaly detection framework that integrates a domain-invariant feature extractor with an unsupervised anomaly detection module. Leveraging the pre-trained foundation model MOMENT, the extractor employs Random Forest Classifiers to disentangle embeddings into machine-related and condition-related features, with the latter serving as representations which are invariant to differences between individual machines. These refined features enable the downstream anomaly detectors to generalize effectively to unseen target machines. Experiments on an industrial dataset collected from three different machines performing nominally the same operation demonstrate that the proposed approach outperforms both the raw-signal-based and MOMENT-embedding feature baselines, confirming its effectiveness in enhancing cross-machine generalization.

**Keywords**: anomaly detection, time series, domain generalization, out-of-distribution generalization, unsupervised learning

# 1. Introduction

With the increasing demand for product quality and manufacturing system resilience, efficient data-driven anomaly detection methods have become increasingly important [1]. The failure to detect even minor deviations from normal operating conditions may result in significant quality and financial losses due to production downtime and product defects, especially in highly sophisticated and sensitive manufacturing areas, such as semiconductor or pharmaceutical manufacturing. Manufacturing machines in modern industrial systems are equipped with a large number of interconnected sensors, which provides an opportunity to develop efficient data-driven anomaly detection approaches based on the readings produced by those sensors [2].

Most existing multivariate time-series based anomaly detection methods assume that the observed training data and the unobserved testing data follow the same distributions [3]. However, this assumption often fails in practice, leading to the so-called *out-of-distribution* (OOD) issues. Machines of the same type may exhibit different behavior in both normal and abnormal scenarios due to various unobservable and untraceable factors, such as external temperature fluctuations, humidity variations, external vibrations, and machine settings. A common approach to mitigate the OOD issue is to collect data from the target domain and adapt the model trained on the source domain to the target domain—an approach known as domain adaptation (DA) [4]. Unfortunately, obtaining target-domain data is often difficult or even infeasible in many real-world applications [5]. Moreover, collecting data from all possible domains, even when target-domain data are available, requires substantial time and labor investment.

The aforementioned OOD challenge and the absence of target-domain data can be addressed by *domain generalization* (DG) techniques [3]. The objective of DG approaches is to train models on data from a limited number of source domains such that they can generalize effectively to unseen target domains. Although DG has achieved remarkable progress in various fields and has attracted increasing research attention, most studies have focused on computer vision and image processing realms [6]-[8]. In contrast, research on time-series domain generalization (TS-DG) remains relatively nascent, particularly in the context of manufacturing applications [9].

The central idea of most DG approaches lies in extracting domain-invariant features that are representative of system behavior across different source domains, thereby enabling robustness to unseen target domains and improving downstream predictive performance [10]. There exist multiple strategies for extracting such domain-invariant features. One major line of work falls under domain alignment, where the differences among source-domain distributions are minimized using various statistical distance measures. Depending on the assumed domain shift, the alignment can be performed on the input feature distribution [11][12], or on the conditional label distribution given the feature space [13]. Alternatively, domain-invariant representations can also be learned without explicit distribution matching by directly reducing domain discrepancies in the latent feature space [14][15]. However, enforcing complete invariance across all features or model components can be challenging, especially as the number of source domains increases. To address this limitation, several studies aim to learn *disentangled representations*, allowing part of the learned representation to remain domain-specific [16][17]. One intuitive approach is to design a model architecture that explicitly decomposes representations into domain-specific and domain-invariant components, based on which domain-invariant representation can be obtained by applying a domain-invariant binary mask to the feature vectors [18]. In [19] and [20], generative modeling is employed to achieve feature disentanglement, where different modules are trained in an adversarial way to capture domain-specific and object-related information. In both cases, the disentanglement process typically requires access to domain labels.

Overall, efficient extraction of domain-invariant features is the cornerstone of TS-DG. However, implementing these methods is non-trivial, as they often demand sophisticated deep learning architectures and the integration of advanced AI/ML techniques, such as generative modeling, reinforcement learning, and ensemble learning.

The recent development of large time-series models offers a potential "out-of-the-box" solution to the DG challenge [21]. Inspired by the success of Large Language Models (LLMs), researchers have proposed large-scale, pre-trained, and task-agnostic *foundation models* for time-series analysis [22]. One representative example is **MOMENT** [23], which is pre-trained on large-scale, multi-domain datasets and employs an attention-based transformer architecture to capture both short- and long-range temporal dependencies. This large-scale pre-training enables foundation models to exhibit strong domain-generalization capabilities across diverse time-series tasks. Nevertheless, the limitations of such representations are also evident. Namely, since foundation models are trained in a task-agnostic manner and not optimized for any specific dataset or application, their representations may be suboptimal for downstream anomaly detection in real-world industrial systems. Consequently, additional adaptation or feature refinement is usually necessary to fully exploit the potential of foundation model representations in specific practical implementations of sensor time-series based anomaly detection.

Inspired by the concept of feature disentanglement, this paper proposes a cross-machine anomaly detection framework in which a domain-invariant feature extractor leverages the representational power of large time-series foundation models. Since such models have already demonstrated domain generalization capabilities, it is reasonable to assume that part of the representations they produce remains stable under domain shifts. In the proposed domain-invariant feature extractor, Random Forest Classifiers (RFCs) [24] are employed to identify two categories of features within the MOMENT embedding, one being features that are critical to the anomaly detection task, while another being features that primarily capture domain-specific identity information. A domain-invariant mask, derived from the extractor, is then applied to the MOMENT embedding representations, yielding dimensionally reduced, domain-invariant representation. Following this extraction stage, a downstream Unsupervised Anomaly Detection (UAD) [25] model is applied. UAD methods are widely adopted in industrial applications because they eliminate the need for costly and labor-intensive collection of anomalous data. Assuming that anomalies are rare in the training dataset, UAD models learn the distribution of normal data and flag any new samples that deviate significantly from this learned distribution. Consequently, integrating UAD with domain-invariant features enables the proposed framework to remain robust when encountering unseen anomalous samples from unseen target domains.

The details of the proposed methodology are presented in the remainder of this paper, which is organized as follows. Section 2 reviews the related state-of-the-art work, while Section 3 introduces the newly proposed domain-invariant feature extractor within the cross-machine anomaly detection framework. Section 4 describes the industrial dataset consisting of sensor readings from multiple machines which was used to evaluate the proposed anomaly detection framework. Results of this evaluation are presented in Section 5, while Section 6 concludes the paper with a summary of findings and a discussion of future research directions.

## 2. Related Work

### 2.1. MOMENT Model

MOMENT is a family of open-source large-scale pre-trained time-series models [23], which can serve as a foundational component for various time-series analysis tasks. Specifically, one of the functions MOMENT can do is to generate embeddings of length 1024 from multivariate time series data by its encoder. The encoder component of the MOMENT model employs a multi-head attention transformer structure aimed at capturing intricate long-range temporal dynamics, as well as complex inter-channel relationships [26]. To further enhance representation learning, the MOMENT encoder undergoes masked time series prediction pre-training, which is a widely used self-supervised learning strategy. In addition, its robustness is strengthened by the fact that MOMENT is trained on an extensive and diverse collection of publicly available datasets across multiple domains, including healthcare, engineering, and finance. This broad training corpus enables the model to generalize well across time series with varying temporal resolutions, numbers of channels, sequence lengths, and amplitudes. As a result, MOMENT is particularly well-suited for general-purpose time-series analysis, which is why it was adopted in this study.

### 2.2. Unsupervised Anomaly Detection

Classical unsupervised anomaly detection methods have been widely explored for equipment condition monitoring and sensor-based fault detection. In this study, we employed four kinds of anomaly detectors: Support Vector Data Description (SVDD), Isolation Forest, Autoencoder, and GANomaly.

**Support Vector Data Description (SVDD)** [27] is a classification and anomaly detection technique inspired by the Support Vector Machine (SVM) algorithm. SVDD constructs a minimal hypersphere which encloses all, or most of the training data in the feature space. Thus, the squared distance between the encoded feature vector of a sample time-series and the hypersphere center can serve as an anomaly score, with a higher score indicating a longer distance and thus a higher probability of anomaly. This method can be kernelized to capture nonlinear relations, making it effective for detecting subtle deviations in high-dimensional sensor data. However, its performance is often sensitive to the choice of kernel and feature scaling [28].

**Isolation Forest** [29] is an ensemble-based unsupervised anomaly detector, where a set of binary trees is built by repeatedly splitting the data space based on randomly chosen features. Under the assumption that the anomalies are rare and significantly different from the majority of normal samples, they are expected to require fewer splits and thus shorter path lengths to be isolated in a leaf node. Therefore, the anomaly score for each sample is defined by its average path length across all trees, with a shorter path length indicating a higher likelihood of anomalous behavior. Isolation Forest is computationally efficient, non-parametric, and scales well to large datasets, making it suitable for real-time monitoring [29].

**Autoencoders** [30] are a family of neural networks trained to reconstruct its input through a low-dimensional latent space. By learning from predominantly normal data, the encoder–decoder pair learns to represent the nominal distribution, while anomalous samples—unseen or rare during training—yield larger reconstruction errors. This reconstruction-based strategy has been widely applied in anomaly detection practice due to its ability to capture nonlinear and non-Gaussian patterns, with a straightforward anomaly scoring based on the reconstruction loss.

**GANomaly** [31] model is an unsupervised anomaly detection framework which extends the traditional Autoencoder concept by embedding it within a Generative Adversarial Network (GAN) architecture. Specifically, GANomaly introduces an encoder–decoder–encoder structure, where the first encoder–decoder pair reconstructs input samples, while the second encoder maps the reconstructed samples back into the latent space. A discriminator is trained in the adversarial manner to encourage the generator to produce reconstructions that are as indistinguishable as possible from real inputs. The anomaly score is defined as the distance between the latent representations of the original and reconstructed samples. By learning the latent distribution of normal data, the model can effectively identify instances that deviate significantly from this distribution as anomalies. Although this adversarial formulation enhances representation quality and improves the distinction between normal and abnormal behaviors, it can also introduce instability during training due to the inherent challenges of GAN optimization [32]

# 3. Methodology

This study addresses cross-machine anomaly detection, where knowledge extracted using solely data from one set of machines is transferred to accomplishing anomaly detection on an unseen target machine. Formally, we observe source datasets $\mathcal{D}_S = \{\mathcal{D}_{S_1}, \mathcal{D}_{S_2}, \dots, \mathcal{D}_{S_m}\}$ and an unlabeled target-domain dataset $\mathcal{D}_T$. Each source-domain dataset $\mathcal{D}_{S_k} = \{(\mathbf{X}, y)_{S_k}^i\}_{i=1}^{N_{S_k}}$ contains $N_{S_k}$ labeled samples, consisting of an input $\mathbf{X} = (x_1, x_2, \dots, x_c) \in \mathbb{R}^{n_t \times c}$ and the corresponding label $y$, with input $\mathbf{X}$ represents a single signal record composed of signals from $c$ sensors over $n_t$ time steps and the condition label $y = 0$ indicating the normal machine operation, while label $y = 1$ indicates abnormal conditions. The target-domain dataset is structured in the same way, but without labels. Although all machines are assumed to be nominally identical and performing the same task, their sensor signals are allowed to differ due to practical factors, such as variations in usage, maintenance histories, manufacturing variations, calibration offsets, initial setup variations and environmental conditions.

This section first presents the construction of the domain-invariant feature extractor leveraging the embedding layer of the pre-trained MOMENT model. Then, the domain-invariant features will be integrated to the downstream anomaly detection framework.

## 3.1. Domain-Invariant Features from Pretrained MOMENT Model

Using the MOMENT model, each signal record can be transformed into an embedding representation $e \in \mathbb{R}^{1024}$. Based on these embeddings, we construct the source-domain datasets $\mathcal{D}_{S_k}^e = \{(e_i^{(S_k)}, y_i^{(S_k)})\}_{i=1,2,\dots,N_{S_k}}$, which contain the embedding representations $e_i^{(S_k)}$ and corresponding labels $y_i^{(S_k)}$ of all samples in the original source-domain datasets $\mathcal{D}_{S_k}$. Similarly, the target-domain embedding dataset is defined as $\mathcal{D}_T^e = \{e_i^T\}_{i=1,2,\dots,N_T}$, which consists only of the embedding representations $e_i^T$ for the target-domain samples, without the corresponding labels.

Robust domain generalization relies on learning domain-invariant features. Ideally these features should strongly correlate with the machine condition (normal vs. abnormal), while showing minimal correlation with machine identity characteristics. To identify such features, we employ a Random Forest Classifier (RFC) [24], which has been widely used for feature importance estimation and selection. Specifically, RFC quantifies feature importances by measuring the total reduction in Gini impurity [33] that each feature contributes during classification.

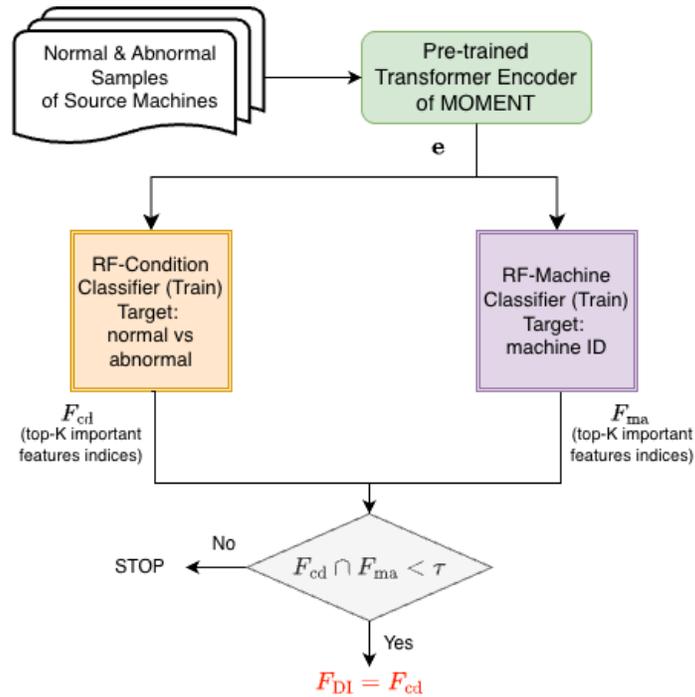

*Figure 1. Domain-invariant feature extractor. First, samples (signal records) from different machines are embedded using the MOMENT model. Next, two random forest classifiers (RFCs) are trained: one for classifying machine condition (RF-Condition) and the other for identifying the machine source (RF-Machine). If the overlap between the important feature sets identified by the two classifiers is smaller than a predefined threshold, the features from RF-Condition are regarded as domain-invariant features.*

Figure 1 illustrates the procedure for identifying domain-invariant features. We hypothesize that among the 1024 embedding features, some features are critical for distinguishing machine condition, whereas others contribute little to the task. Therefore, we train RFCs on the source-domain embedding dataset $\mathcal{D}_S^e$ for two distinct classification tasks:

1. **Machine Condition Classification,** which identifies normal versus abnormal machine states. The top $N_I$ most important features for this task constitute the feature set $F_{cd}$.
2. **Machine Identity Classification,** which determines from which source machine an embedding originates. The top $N_I$ most important features for this task constitute the feature set $F_{ma}$.

If the overlap between $F_{cd}$ and $F_{ma}$ is small (e.g., less than 10% of $N_I$), then $F_{cd}$ is considered domain-invariant because it predominantly relates to machine condition rather than the machine identity. Conversely, a substantial overlap would suggest the proposed approach is unsuitable.

If the overlay task is passed, we define the domain-invariant feature index set as $F_{DI} = F_{cd}$. Accordingly, the domain-invariant representation $e_{DI}$ of each signal record $(\mathbf{X}, y)$ is obtained by retaining only the features indexed by $F_{DI}$ from the full MOMENT embedding $e$.

### 3.2. Cross-machine Anomaly Detection Framework

Figure 2 illustrates the proposed cross-machine anomaly detection framework, where a domain-invariant feature extractor is employed to construct the domain-invariant representations. The downstream anomaly detection module can adopt any unsupervised method, such as Deep SVDD, Isolation Forest, Autoencoder, or GANomaly. In the training stage, only normal samples from the source machines ($\mathcal{D}_S$) are used to model the distribution of normal machine behavior. By learning consistent normal patterns across source domains, the framework detects anomalies as samples that deviate from this learned normal behavior distribution irrespectively of the source machine, thereby eliminating the need for explicit domain adaptation on the unlabeled target machine.

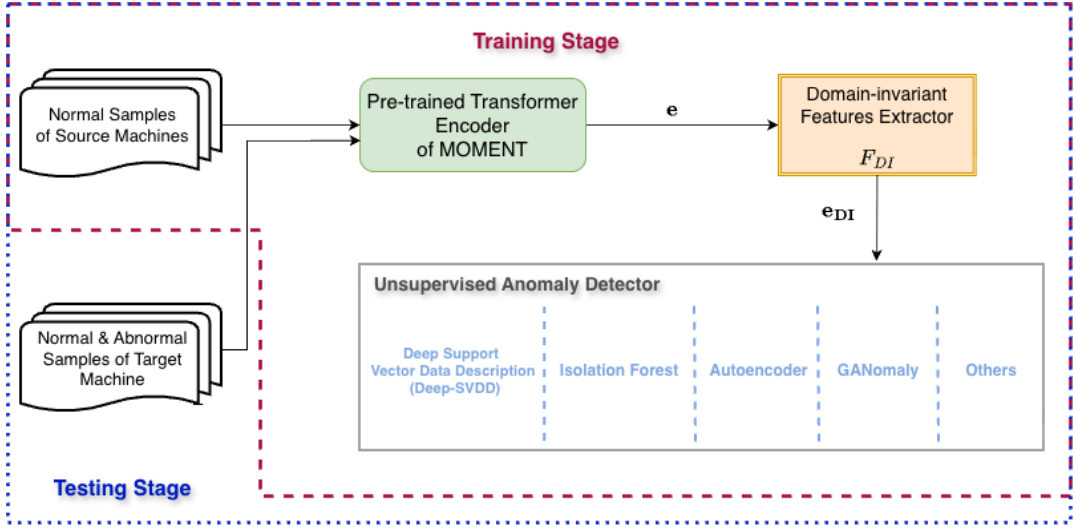

*Figure 2. Overview of the cross-machine anomaly detection framework. A domain-invariant feature extractor is first applied to obtain domain-invariant representations for each input sample. These representations are then fed into a downstream unsupervised anomaly detection model. During training, only normal samples from the source machines are used.*

# 4. Experiment Setup

## 4.1. Dataset

The proposed methodology was evaluated using data collected from in-house testing equipment that transports loads via a motor-driven conveyor belt system. During each operational cycle, angular torque signals from the motor and angular velocity signals from the load were simultaneously recorded at a high sampling rate. Data from three distinct machines, denoted as $M_1$, $M_2$, and $M_3$, are used in this study. Machines $M_2$ and $M_3$ constitute the source domain, represented as

$$\mathcal{D}_S = \{\mathcal{D}_{M_2}, \mathcal{D}_{M_3}\} = \left\{\left\{\left(\mathbf{X}_i^{(M_2)}, y_i^{(M_2)}\right)\right\}_{i=1,2,\ldots,11000}, \left\{\left(\mathbf{X}_i^{(M_3)}, y_i^{(M_3)}\right)\right\}_{i=1,2,\ldots,11000}\right\}$$

where each machine provides 10,000 normal and 1,000 abnormal signal records. Machine $M_1$ forms the target domain $\mathcal{D}_T = \left\{\mathbf{X}_{M_1}^i\right\}_{i=1,2,\ldots,1000}^{1,000}$, which includes 900 normal and 100 abnormal signal records.

It is important to note that the causes of anomalous behavior differ among the three machines. For instance, the anomalous behavior in one machine might result from wear of components, whereas the anomalies in other machines may be caused by different root causes. Consequently, the anomalies in the target domain (M$_1$) are unseen to the model trained on M$_2$ and M$_3$.

### 4.1.1. Dataset for Domain-Invariant Feature Extraction

Domain-invariant feature extraction exclusively relied on the source domain dataset, with the target domain data not being available during training. Specifically, the 2,000 abnormal samples and 2,000 randomly picked normal samples in $\mathcal{D}_S$ are shuffled and partitioned in such a way that 80% of samples formed the training set, while the remaining 20% formed the testing set for the Random Forest Classifiers (RFCs). An equal number of normal and abnormal samples are chosen to maintain the class balance, thereby enhancing the reliability and accuracy of RFCs.

### 4.1.2. Dataset for Anomaly Detection

All normal samples in the source domain dataset $\mathcal{D}_S$ constitute the training set, and all samples from target domain dataset $\mathcal{D}_T$ form the testing set. To enhance the diversity of the training set, two data augmentation methods are applied to training set:

1. **Time-shifting augmentation**: Normal signals are shifted forward by five time steps, generating temporally altered but label-preserving variants.
2. **Mix-up augmentation**: New signal records are synthesized by linearly combining existing ones. Given two time-series samples $x^1$ and $x^2$, a new sample $x'$ is created as: $x' = \lambda x^1 + (1 - \lambda)x^2, \lambda \in [0,1]$.

Time-shifting augmentation simulates observations at different operational starting points while maintaining signal alignment. The Mix-up augmentation efficiently generates transitional variations between machines' data, thus acting as a regularizer and reducing model sensitivity to idiosyncrasies related with machine identity. In this study, 10,000 time-shifted samples are generated based on the M2's normal samples. And 17,000 mixed samples are generated by combining signals from M2 and M3. Half of these mixed samples use equidistantly sampled $\lambda$ values between 0.1 and 0.3, mimicking signals more similar to M3, while the remaining half use $\lambda$ values between 0.7 and 0.9, mimicking signals more similar to M2.

In summary, the training set comprises 47,000 normal samples from M1 and M2, including 20,000 original signals, 10,000 time-shifted signals, and 17,000 mix-up-augmented signals. The testing set contains 900 normal and 100 abnormal samples from M1. Besides, a validation set $\mathcal{D}_{val}$ is formed by randomly sampling 10% of the training set and is used to tune model hyperparameters.

## 4.2. Hyperparameter Tuning

The dimensionality of domain-invariant features, $N_I$, is chosen to maximize the dimensionality of domain-invariant representation, while keeping the overlap ratio of $F_{cd}$ and $F_{ma}$ under 10%. Hyperparameters for RFCs in the domain-invariant feature extractor are selected via a full grid search to maximize the F1 score. However, for the unsupervised anomaly detectors, training does not involve labeled data, rendering the F1 score unavailable. Although abnormal sample labels exist in the source domain datasets which have been used for domain-invariant feature extraction, these abnormal samples are deliberately excluded from forming the validation set. One reason for this exclusion is that the domain-invariant extractor and unsupervised anomaly detector can be trained independently on datasets obtained from different machines. More importantly, the objective of the proposed anomaly detection framework is to detect any abnormal condition, including previously unseen anomalies, rather than overfitting to specific known failure types present in the validation set.

Therefore, to tune the hyperparameters of the unsupervised anomaly detector, we employ the Expected Anomaly Gap (EAG) metric [34]. EAG is a label-free measure quantifying how effectively a detector distinguishes the extreme tail of its normal data score distribution from its bulk, with a higher EAG score indicating better discrimination capability. In this study, EAG is computed using the anomaly score distribution of the training set, focusing on the upper tail region between the 90th and 100th percentiles. More detailed description of the concept of EAG and methods to compute it can be found in [34].

Hyperparameters of the Deep-SVDD, Isolation Forest and Autoencoder methods are tuned through a grid search based on EAG scores. For the **Deep-SVDD**, the search space included hidden dense layer sizes $\{[128,64,16],[256,128,32]\}$, training epochs $\{50,100\}$, batch sizes $\{128,256\}$, $L_2$ regularization coefficients $\{10^{-4}, 10^{-3}\}$, and dropout rates $\{0.0,0.2\}$. For the **Isolation Forest**, we evaluated the number of estimators $\{100,200,400\}$, maximum samples $\{256,512\}$, maximum feature fractions $\{1.0,0.8,0.6\}$, and bootstrap options $\{False,True\}$. For the **Autoencoder**, we varied the latent dimension $\{8,16,32\}$, dense encoder–decoder layer sizes

$\{(128,64),(256,128)\}$ and $\{(64,128),(128,256)\}$, dropout rates $\{0.0, 0.1\}$, learning rates $\{10^{-3}, 5 \times 10^{-4}\}$, and training epochs $\{50, 100\}$. For each model, the configuration yielding the highest training-set EAG score was selected for subsequent evaluation on the test data. The architecture of **GANomaly** follows the original GANomaly framework [31], within which we replaced the 2D convolutional layers with 1D convolutional layers, which are better suited for processing time-series signals. All other hyperparameters of the GANomaly are adopted from the original implementation [31], except that the dimension of latent representations $z$ is chosen via a grid search maximizing the EAG score over the set $\{30, 40, 50, 60\}$.

Due to the commonly observed overfitting in adversarial-based models [35], we also tuned the early-stopping criterion for the GANomaly method. The metric of Validation Error $R$ is introduced the average $\mathcal{L}_1$ distance between each input and its reconstruction within the validation set $\mathcal{D}_{val}$. This metric was calculated every 100 batches, and the training was stopped once the change between consecutive evaluations, $|R^{(k-1)} - R^{(k)}|$, was below a threshold (tolerance) for a specified number (patience) of consecutive evaluations. Both the *tolerance* and *patience* parameters were tuned via a grid search maximizing the EAG score, with *tolerance* values taken from the set $\{10^{-3}, 5 \times 10^{-4}, 10^{-4}\}$ and *patience* values taken from the set $\{3, 5\}$.

## 5. Results

The performance of experimental results is reported using precision, recall, F1-score, Area Under the Receiver Operating Characteristic Curve (AUC-ROC), and Area Under Precision-Recall Curve (AUPRC) [36]. Precision, recall, and F1-score are defined respectively as

$$\text{Precision} = \frac{TP}{TP + FP}, \quad \text{Recall} = \frac{TP}{TP + FN}, \quad \text{F1 score} = \frac{2 \cdot \text{Precision} \cdot \text{Recall}}{\text{Precision} + \text{Recall}} \quad (1)$$

where TP, FP, and FN denote the number of true positives, false positives, and false negatives, respectively. In contrast to these threshold-dependent measures, The AUC-ROC and AUPRC are threshold-free metrics that provide a holistic evaluation of model performance across all possible decision thresholds. Specifically, the AUC-ROC characterizes the trade-off between true positive and false positive rates, offering a general view of discrimination ability. The AUPRC, on the other hand, focuses on the trade-off between precision and recall, which is particularly informative in highly imbalanced datasets.

Figure 3 illustrates the overlap between feature sets $F_{cd}$ and $F_{ma}$, showing that only 9 out of 100 features are common between both sets. The random forest classifiers RF-Condition and RF-Machine which were used for these tasks are each trained and evaluated ten times with different random seeds. Their average F-scores were 0.9432 and 0.9863, respectively, demonstrating the reliability of the classification results. Given the small amount of overlap, the most important features identified by RF-Condition ($F_{cd}$) are not confounded by machine identity and can therefore be regarded as the domain-invariant feature set $F_{DI}$.

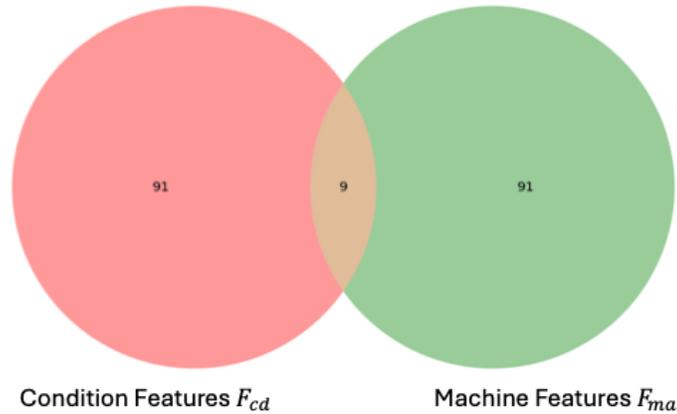

*Figure 3. The overlap between important condition feature set $F_{cd}$ and important machine identity feature set $F_{ma}$*

Figure 4 presents the AUC-ROC and AUPRC results for various anomaly detectors evaluated under three input regimes: (i) the raw signal records, (ii) MOMENT embedding representations and (iii) the domain-invariant feature representations $e_{DI}$ introduced in Section 3.1. Each value is averaged over ten runs with different random seeds, reported along with its standard deviation and displayed as error bars. GANomaly fails on the MOMENT embeddings, possibly due to the higher dimensionality of that embedding and the fragility of adversarial training of GANomaly. However, GANomaly can be trained steadily on the domain-invariant features. Accordingly, the domain-invariant feature extraction also shows the ability in feature reduction, making it easier to be applied for multiple different types of anomaly detection methods.

The domain-invariant features extracted via proposed domain-invariant feature extractor show clear superiority for domain generalization. All the anomaly detectors attain their highest AUC-ROC and AUPRC on domain-invariant features, compared with the ones on raw signals or MOMENT embeddings. The only exception is Autoencoder, whose AUC-ROC on domain-invariant features is marginally lower (by 0.12) than that of the MOMENT embeddings.

One can also note that the tree-based ensemble Isolation Forest exhibits the largest performance gain and achieves the highest AUC-ROC and AUPRC on domain-invariant features. A plausible explanation is that these features are ranked and filtered by a Random Forest Classifier, which is also a tree-based ensemble relying on axis-aligned threshold splits. Therefore, the features identified as important by Random Forest Classifier tend to induce large variance reduction or clear boundaries upon splitting, which particularly benefits the Isolation Forest. In contrast, other representation-learning-based methods rely on continuous latent representations and nonlinear manifolds rather than axis-aligned separations.

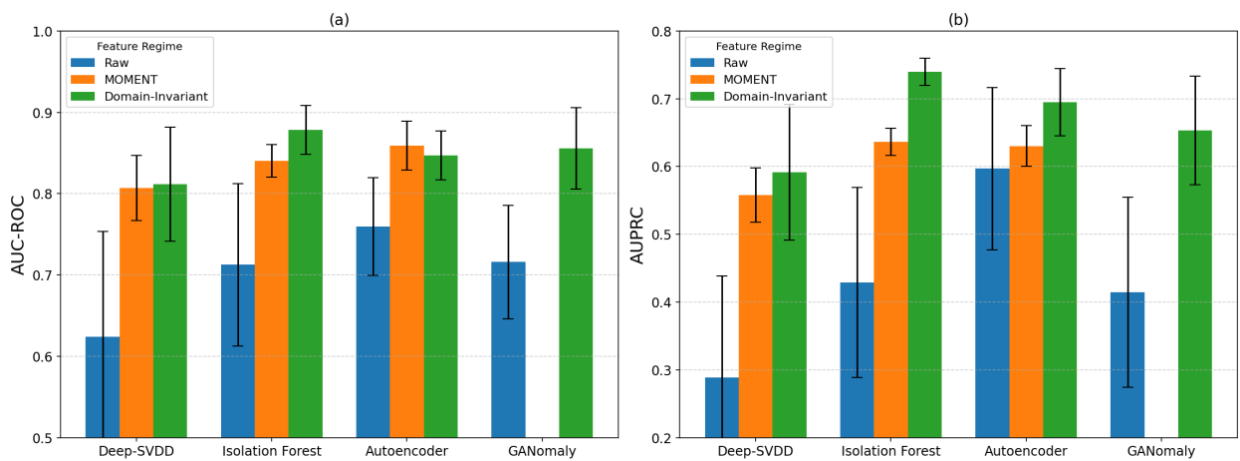

Figure 4. AUC-ROC (a) and AUPRC (b) across feature regimes for various anomaly detection models. The values for GANomaly with MOMENT embedding features are unavailable due to failure of training.

Figure 5 presents the variation of the average F1-score as the anomaly threshold changes from the 90th to the 100th percentile of training-set anomaly scores. In Figure 5(b) the negative average path length across trees is used as the anomaly score for Isolation Forest.

Across most anomaly detectors, the domain-invariant features raise the F1 over a wide ranged anomaly threshold selection. Isolation Forest exhibits the strongest and most stable gains: its F1-scores obtained from domain-invariant features are between 0.6 and 0.7 across the percentile range, consistently outperforming MOMENT embeddings. Deep-SVDD also benefits from domain-invariant features up to roughly the 94th percentile, though F1-scores yielded by MOMENT embeddings become more stable at higher thresholds. Autoencoder, however, struggles to generalize under any feature representation. By contrast, GANomaly—which integrates the Autoencoder within an adversarial learning framework and computes anomaly scores in the latent space—shows a marked and consistent improvement when using domain-invariant features. Besides, it should be noted that using raw signal records as features consistently fails to generalize across domains in any anomaly detector. The resulting F1-scores remain around 0.18, which corresponds to the degenerate case where nearly all test samples are predicted as anomalies.

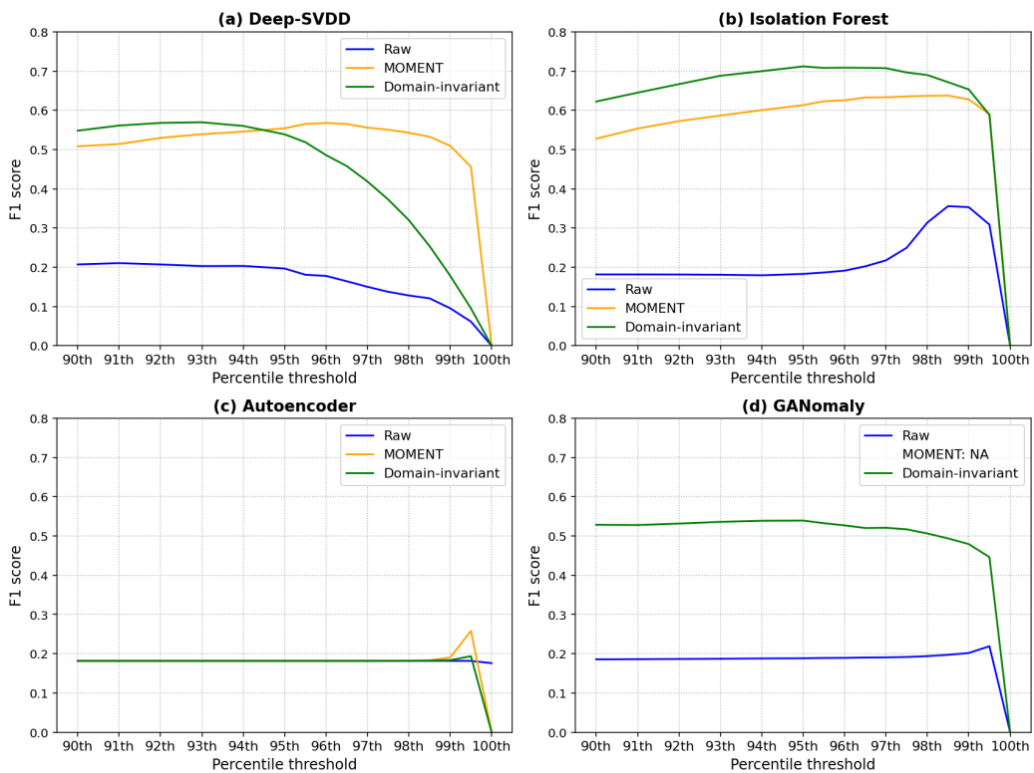

*Figure 5. Average F1-score over ten independent runs as a function of the anomaly-score threshold, varied from the 90th to the 100th percentile of the training-set distribution in increments of 0.5 percentile, for (a) Deep-SVDD, (b) Isolation Forest, (c) Autoencoder, and (d) GANomaly.*

# 6. Conclusion and Future Work

This paper presents a cross-machine time-series anomaly detection framework that integrates a newly proposed domain-invariant feature extractor with the downstream unsupervised anomaly detector. The feature extractor employs two Random Forest Classifiers (RFCs) to disentangle the features derived from the MOMENT embeddings into two components: (i) features which are strongly correlated with machine operating condition regardless of the specific machine, and (ii) features which strongly related with machine identity. When the overlap between these two feature groups is small enough, the condition-related features can be regarded as domain-invariant features, enabling more robust and transferable anomaly detection across machines.

Experimental results involving multiple individual machines demonstrate that the proposed domain-invariant feature regime consistently outperforms both the raw-signal and MOMENT-embedding regimes, achieving higher performance in terms of AUC-ROC and AUPRC, with the only exception being a marginal 0.14% decrease compared to the MOMENT embedding in the Autoencoder anomaly detector. Among the tested anomaly detectors, the tree-based Isolation Forest exhibited the greatest improvement by using domain-invariant features, compared to encoding-based and reconstruction-based approaches. Furthermore, the reduced dimensionality of the domain-invariant features contributes to more stable training behavior in adversarial learning methods, such as GANomaly.

Several directions can extend this work in the future studies. Although the current framework benefits from the representational power of large time-series foundation models, their generality implies that embeddings are not always effective for specific datasets or tasks. Continued advancement in the development of more powerful and task-adaptive time-series foundation models could therefore further enhance the proposed approach. Additionally, the current feature extraction framework assumes a limited overlap between the condition-related and machine-related feature components, which may not always hold in practice. Future research could address this limitation by exploring more sophisticated feature disentanglement strategies, such as incorporating adversarial learning mechanisms to achieve finer separation between the condition-related and machine-related features.

## Acknowledgment

This work was supported in part by Tokyo Electron Ltd. through a sponsored research agreement with the University of Texas at Austin. The authors would like to thank the Tokyo Electron Ltd for their generous support, advice, and plethora of data. Any opinions, findings, conclusions, or recommendations expressed in this paper are those of the authors and do not necessarily reflect the views of Tokyo Electron Ltd.

## Data Availability

The participants of this study did not give written consent for their data to be shared publicly, so due to the sensitive nature of the research supporting data is not available.

## Disclosure of Interest

There are no relevant financial or non-financial competing interests to report.